# Dynamic simulation of task constrained of a rigid-flexible manipulator

**Atef A. Ata[1] & Habib Johar[2]**
[1] International Islamic University Malaysia, Department of Mechatronics Engineering,
atef@iiu.edu.my
[2] International Islamic University Malaysia, Department. of Mechatronics Engineering,
habib3133@hotmail.com

*Abstract: A rigid-flexible manipulator may be assigned tasks in a moving environment where the winds or vibrations affect the position and/or orientation of surface of operation. Consequently, losses of the contact and perhaps degradation of the performance may occur as references are changed. When the environment is moving, knowledge of the angle α between the contact surface and the horizontal is required at every instant. In this paper, different profiles for the time varying angle α are proposed to investigate the effect of this change into the contact force and the joint torques of a rigid-flexible manipulator. The coefficients of the equation of the proposed rotating surface are changing with time to determine the new X and Y coordinates of the moving surface as the surface rotates.*
*Keywords: Constrained motion; Rigid-flexible manipulator; Moving environment; Analytical solution, Profiles.*

### 1. Introduction

In the last two decades, robotic constrained motion applications, where the end-effector of the robot is always in contact with its environment, have received a considerable attention. Typical examples of such applications include grinding, deburring, cutting, polishing, drilling, fastening, etc. To execute these tasks successfully it is necessary to control both motion of the robot and the contact force between the end-effector and the environment. Despite the voluminous research done in the last decade, the study of dynamics and control of the constrained motion of the flexible manipulators remains open for further investigations.

Modeling, simulation, and control of manipulators in constrained motion attract many researchers for the last two decades. In 1981, the first attempt in dealing with this problem for rigid manipulators was proposed by applying hybrid position/forcce control scheme ( Raibert, M. & Craig, J., 1981). For rigid-flexible manipulator, the work by (Su, et al. 1990) represented the earliest study in force control of constrained maneuver. In 1994, a dynamic modeling and force/motion control of constrained flexible robot was presented ( Ulsoy, G. & Hu, F., 1994). In their model, the flexible link has an additional degree-of-freedom to move axially into the first rigid link. The dynamic hybrid position/force control of a rigid-flexible (Matsuno, F. &Yamamoto, K., 1994) and two-link flexible manipulator ( Matsuno, et al, 1994) where the elastic deformations were approximated by B-spline functions were Investigated. The inverse force and motion control of constrained three- axis elastic robot based on non-linear inversion and stabilization was presented (Yim, W. & Singh, H., 1995). Using a two-time scale force/position controller, a general systemic model of flexible robot interacting with a rigid environment was addressed (Rocco, P. & Book, W., 1996). The recent work by  (Shi, et al, 1999)], A mathematical model of a constrained rigid-flexible manipulator based on Hamilton's principle was derived. A multi variable controller was proposed for the simultaneous motion and force control was also investigated. A parallel force and position control scheme of flexible manipulators using perturbation theory was proposed ( Siciliano, B. & Villani, L., 2000). In 2000, an algorithm employing the computed torque method in the free space and hybrid motion/force controller for rigid manipulators was presented (Bennon, et al., 2000). Experimental results on non-moving and moving environment were also illustrated. While in 2001, a position and force control scheme for a class of flexible joint robots during constrained motion tasks with model uncertainties based on singular perturbation theory was developed (Hu, Y. & Vukovich, G. 2001).
Dynamic modeling and simulation of constrained motion of rigid-flexible manipulator in contact with a compliant surface was addressed (Ata, A. & and Ghazy, S., 2001). While in 2003, they continued their work towards finding an optimal motion trajectory for the constrained motion based on the minimum energy consumption (Ata, et al., 2003). In this study, an extension to the results in the previous work is presented in such away to further study the interaction of the joint motion profile and the angular motion profile of the constraint surface. The constrained surface is assumed to be rotating, therefore, different angular profiles for the surface are proposed to investigate the effect of this change into the constrained force and the joint torques.

## 2. Dynamic Modeling

Consider the two-link planar manipulator shown in Figure 1. The first link is rigid and the second link is assumed to be flexible. The end-effector of the robot is in contact with a moving environment during the task. The tip payload consists of mass $m_3$ with moment of inertia $I_3$ about its own axis of rotation. The flexible link is assumed to be thin and slender so it can be modeled as an Euler-Bernoulli's beam of length $l_2$, uniform mass density $\rho_2$, the cross section $I_2$. Applying the Virtual Link Coordinate System (VLCS), consider $\theta_2(t)$ be the angle of a line pointing from the second joint to the tip mass as described in reference (Benati, M. & Morro, A., 1994). The transversal bending deflection $w(x,t)$ at a point $x\varepsilon[0, l_2]$ along the second link is described with respect to the virtual link. Longitudinal deformations are neglected. No damping is assumed and the manipulator moves in the horizontal plane so, the gravity is not considered.

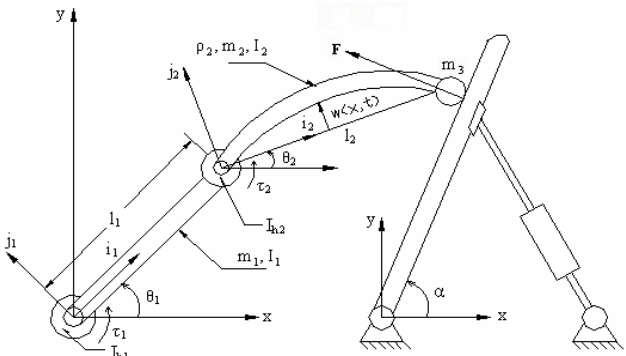

Fig. 1: A two link rigid-flexible manipulator with tip mass

The equations of motion can be derived using the extended Hamilton's principle (Ata, et al., 2003). The constrained motion of the rigid-flexible can be described as:

$$\tau = M(\theta)\ddot{\theta} + V(\theta,\dot{\theta}) + \tau_r + \tau_f \qquad (1)$$

where:

$$\tau = \begin{bmatrix} \tau_1 \\ \tau_2 \end{bmatrix}$$

in which $\tau_1$ and $\tau_2$ are the rigid and flexible torques respectively.
$M(\theta)$ is a 2 by 2 inertia matrix of the manipulator and its elements can be described by:

$$M_{11} = (I_1 + I_{h1} + m_2 l_1^2 + m_3 l_1^2)$$
$$M_{12} = (m_2 l_1 \bar{x}_2 + m_3 l_1 l_2)\cos(\theta_2 - \theta_1)$$
$$- \int_0^{l_2} \rho l_1 w \sin(\theta_2 - \theta_1) dx$$
$$M_{21} = M_{12}$$
$$M_{22} = (I_2 + I_{h2} + m_3 l_2^2) + \int_0^{l_2} \rho w^2 dx$$

$V(\theta,\dot{\theta})$ is the Coriolis and centripetal terms and can be given by the column vector:

$$V(\theta,\dot{\theta}) = \begin{pmatrix} v_{11} \\ v_{21} \end{pmatrix}$$

$$v_{11} = -(m_2 l_1 \bar{x}_2 + m_3 l_1 l_2)\dot{\theta}_2^2 \sin(\theta_2 - \theta_1)$$
$$+ \int_0^{l_2} \rho_2 l_1 [(\ddot{w} - w\dot{\theta}_2^2)\cos(\theta_2 - \theta_1) - (2\dot{w}\dot{\theta}_2)\sin(\theta_2 - \theta_1)] dx$$

$$v_{21} = (m_2 l_1 \bar{x}_2 + m_3 l_1 l_2)\dot{\theta}_1^2 \sin(\theta_2 - \theta_1)$$
$$+ \int_0^{l_2} \rho_2 \{-2w\dot{w}\dot{\theta}_2 + x\ddot{w} + l_1 w\dot{\theta}_1^2 \cos(\theta_2 - \theta_1)\} dx$$

$\tau_f$ is the torque due to friction. Frictional force is a highly nonlinear phenomenon that is difficult to model accurately. In the present study, we neglect the effect of this friction force.
$\tau_r$ is the interaction torque due to the contact with the environment and given as:

$$\tau_r = J^T F = J^T \begin{bmatrix} F_x \\ F_y \end{bmatrix}$$

where $J^T$ is the transpose Jacobian of the system.

$$J^T = \begin{pmatrix} -l_1 \sin\theta_1 - l_2 \sin\theta_2 & l_1 \cos\theta_1 + l_2 \cos\theta_2 \\ -l_2 \sin\theta_2 & l_2 \cos\theta_2 \end{pmatrix}$$

$F_x$ and $F_y$ represent the applied forces by the end-effector on the surface in contact including the inertia forces of the end-effector and is modeled with a spring (Su, et al., 1990):

$$F_x = K_s(r_x - l_1\cos\theta_1 - l_2\cos\theta_2) +$$
$$m_3(l_1\ddot\theta_1\sin\theta_1 + l_1\dot\theta_1^2\cos\theta_1 + l_2\ddot\theta_2\sin\theta_2 \qquad (2)$$
$$+ l_2\dot\theta_2^2\cos\theta_2)$$

$$F_y = K_s(r_y - l_1\sin\theta_1 - l_2\sin\theta_2)$$
$$+ m_3(l_1\dot\theta_1^2\sin\theta_1 - l_1\ddot\theta_1\cos\theta_1 \qquad (3)$$
$$+ l_2\dot\theta_2^2\sin\theta_2 - l_2\ddot\theta_2\cos\theta_2)$$

Where $K_s$ is the spring stiffness (N/m), $r_x$, and $r_y$ are the coordinates of the contact point.

For the second link, the equation due to the flexibility effect is given by:

$$\rho_2\{w\dot\theta_2^2 - \ddot w - x\ddot\theta_2 - l_1[\dot\theta_1^2\sin(\theta_2-\theta_1) \qquad (4)$$
$$+ \ddot\theta_1\cos(\theta_2-\theta_1)]\} - k_2 w'''' = 0$$

subject to four boundary conditions.

$$w(0,t) = 0 \qquad (5a)$$
$$w(l_2,t) = 0 \qquad (5b)$$
$$w''(0,t) = -\frac{\tau_{21}(t)}{k_2} \qquad (5c)$$
$$w''(l_2,t) = \omega^2 v'(l_2)I_3(t) \qquad (5d)$$

The fourth boundary condition is due to the existence of the inertia of the tip mass around its own axis of rotation (Clough, R. & Penzien, J., 1993) in which, $\omega$ is the natural frequency of the flexible link, $v'(l_2)$ is an assumed function of spatial coordinate, and $I_3(t)$ is the inertia of tip mass which is a time function because we consider tip mass varies linearly with time.

## 3. Vibration of the flexible link with time-dependent boundary conditions

For an angular displacement $\theta(t)$ and an elastic deflection $w(x,t)$, the total displacement $y(x,t)$ of a point along the flexible link from the hub can be considered as:

$$y(x,t) = w(x,t) + x\theta_2(t) \qquad (6)$$

By ignoring the first term of (4) since its effect is only obvious at very high speed, and substituting from equation (6) into equations (4 and 5) one can get:

$$\rho_2\{-\ddot y - l_1[\dot\theta_1^2\sin(\theta_2-\theta_1) \qquad (7)$$
$$+ \ddot\theta_1\cos(\theta_2-\theta_1)]\} - E_2 I_2 y'''' = 0$$

$$y(0,t) = 0 \qquad (8a)$$
$$y(l_2,t) = l_2\theta_2(t) \qquad (8b)$$
$$y''(0,t) = -\frac{\tau_2(t)}{E_2 I_2} \qquad (8c)$$
$$y''(l_2,t) = \omega^2 v'(l_2)I_3(t) \qquad (8d)$$

Our objective is to find the rigid and flexible hub torques to move the end-effector through a prescribed trajectory by solving the inverse dynamics problem. Solving equations (1 and 4) for the joints torque subject to the boundary conditions (5a-d) to obtain the rigid and flexible hub torques is a very difficult task. This is simply because one has to calculate the elastic deflection of the arm to obtain the flexible hub torque. Unfortunately, the required flexible torque is also included in the time-dependent boundary condition (8c). An alternative approach to the computation of the link deformation is to use approximations for the flexible torque (Asada, et al. 1990). The sequence of calculations to get $\tau_1$ and $\tau_2$, can be summarized as follows:

i- Assume the joint motion profiles $\theta_1(t)$ and $\theta_2(t)$.

ii- Consider $\tau_2$ as a rigid torque without any elastic effect, substitute into equation (8c).

iii- Solve equation (7) subject to the boundary conditions (8) to get $y(x,t)$ and then substitute into (6) to get $w(x,t)$, $\dot w(x,t)$ and $\ddot w(x,t)$

iv- Calculate $F_x$ and $F_y$ from equations (2&3) and substitute into (1) to get $\tau_1$ and $\tau_2$.

For the rigid-flexible manipulator under consideration, spatial discretization techniques (e. g., assumed modes, finite element or Galerkin's method) are typically employed to obtain a finite dimensional system of ordinary differential equation suitable for simulation (Hu and Ulsoy, 1994). The assumed modes method, based on modal expansion, consists in finding a particular solution (admissible function) of the Euler-Bernoulli equation to satisfy only the geometric boundary conditions. On the other hand, the finite element method consists in finding a local solution of the equation over a finite element of the flexible link. In this research we are applying the assumed modes method as a technique for the analytical solution.

The solution for $y(x,t)$ can be obtained using the assumed modes method [Meirovitch, L., 1967 & Low, K., 1989] in the form:

$$y_n(x,t) = \sum_0^n [v_n(x)\zeta_n(t)] + g(x)e(t) \qquad (9)$$
$$+ h(x)f(t) + q(x)p(t)$$

where:
$$e(t) = \theta_2(t) \qquad (10a)$$
$$f(t) = -\frac{\tau_2}{E_2 I_2} \qquad (10b)$$
$$p(t) = \omega^2 v'(l_2)I_3(t) \qquad (10c)$$

The subscript n indicates that the continuous system has been approximated by an n-degree-of-freedom system. The three functions $g(x)$, $h(x)$ and $q(x)$ are of the spatial coordinate alone to satisfy the homogeneous boundary conditions for $v_n(x)$, and $\zeta_n(t)$ is the time function. Taking the lowest possible order of polynomials the functions $g(x)$, $h(x)$ and $q(x)$ can be written as:

$$g(x) = x \tag{11a}$$

$$h(x) = -\frac{1}{3}l_2 x + \frac{1}{2}x^2 - \frac{1}{6l_2}x^3 \tag{11b}$$

$$q(x) = -\frac{1}{6}l_2 x + \frac{1}{6l_2}x^3 \tag{11c}$$

The corresponding eigenfunction $v_n(x)$ can be obtained as [14]:

$$v_n(x) = \sqrt{\frac{2}{\rho l_2}} \sin\frac{n\pi}{l_2}x , \quad n=1,2,3.... \tag{12}$$

The last two nonlinear terms inside the parentheses in equation (6) can be regarded as distributed excitation force with unit density. This effect can be compensated in the time function (Meirovitch, L., 1967) as:

$$\zeta_n(t) = \frac{1}{\omega_n}\int_0^t N_n(\tau)\sin(t-\tau)d\tau \tag{13}$$

where:

$$N_n(t) = N_{n1}(t) - \rho_2 l_1 \int_0^{l_2} v_n(x) \tag{14}$$

$$[\ddot{\theta}_1 \cos(\theta_2 - \theta_1) + \dot{\theta}_1^2 \sin(\theta_2 - \theta_1)]dx$$

$$G_n^* = \int_0^{l_2} v_n(x)g''(x)dx \quad G_n = \int_0^{l_2} v_n(x)\rho_2 g(x)dx$$

$$H_n^* = \int_0^{l_2} v_n(x)h'''(x)dx \quad H_n = \int_0^{l_2} v_n(x)\rho_2 h(x)dx$$

$$Q_n^* = \int_0^{l_2} v_n(x)q'''(x)dx \quad Q_n = \int_0^{l_2} v_n(x)\rho_2 q(x)dx$$

The convolution integral (14) can be evaluated using Duhamel integral method (Clough, R. & Penzien, J., 1993). The simulation algorithm thus developed is implemented in Matlab and the results will be presented in the next section.

**4. Simulation results and discussion**

For the solution of the inverse dynamics problem, three joint motion profiles, namely sine, polynomial, and Gaussian velocity profiles are assumed. In order to investigate the interaction of the two motion profiles on the contact force and the joint torques, four different profiles for the rotation of the constrained surface have been applied to each joint motion profile. These four profiles have in common starting and ending values but they differ in their rates of increase. Accordingly, the angle α of the constrained surface will start from $2\pi/3$ and reach $5\pi/6$ in 5 seconds which is the same time duration as for joint motion. The proposed contact surface is a parabolic shape: $y = ax^2 + bx + c$. The coefficients of the proposed surface are changing with time to determine the new X and Y coordinates of the contact point of the moving surface as the surface rotates. Figures 2 and 3 represent the position of the stationary and rotating surface respectively in X-Y coordinates.

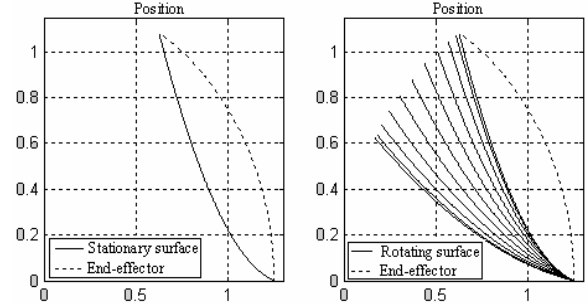

Fig. 2 Stationary surface    Fig. 3 Rotating surface

The system parameters used in this study are given in below:
$l_1$=0.5 m, $l_2$=0.75 m,
$m_1$=0.5 Kg, $m_2= l_2*\rho_2$, $m_3$=0.15+(0.15)t/T
$I_1$=0.0834 Kg.m$^2$, $I_2$=($l_2^3*\rho_2/3$), $I_3=m_3*K^2$ Kg.m$^2$
$I_{h1}$=10*$I_1$, $I_{h2}$=10*$I_2$
$E_2I_2$=2.4507 Nm$^2$

Where T is the time duration of motion and K is the radius of gyration for the tip mass.

The joint torques and the contact force have been simulated first for the stationary surface in Figs 4, 6, 8 to show that they are greatly different from those resulted in Figs 5(a-d), 7(a-d) and 9(a-d) for the rotating surface with different angular velocity profiles.

Simulation plots are presented in groups according to the joint motion profiles. In the Figs 5(a-d), 7(a-d) and 9(a-d), the first word in the Figure's title indicates the angular motion profile of the constrained surface and the second stands for the joint motion profile.

It can be observsed from Figs 5, 7 and 9 that the contact force and the joint torques are certainly affected by the changing of the angular motion profiles for the constrained surface. In all three profiles of joint motion, the contact force tends to increase. While for sinusoidal profile of the constrained surface (Figs. 5a, 7a, 9a), the contact force increases considerably creating uncertainty about the stability the system at the end of the time interval. The polynomial, parabolic and Gaussian profiles of the constrained surface give better distributions for the contact force and joints torque.

On the other hand, the contact force distributions are influenced by the interaction of the angular velocity profile of the environment and the joint motion profile. This can be observed especially in Figs. 9(a-d) where the distributions of the constrained forces are not as smooth as in Figs. 5(a-d) and 7(a-d) due to the Gaussian joint motion profile. As for the joint torques, all four different angular velocity profiles of the constrained surface produce almost same trend with a little bit different peak values in each particular joint motion profile.
In the case of stationary surface, maximum joint torques and contact force result from Gaussian joint motion profile (Fig. 8), followed by sine profile (Fig. 6) and polynomial profile (Fig. 4).

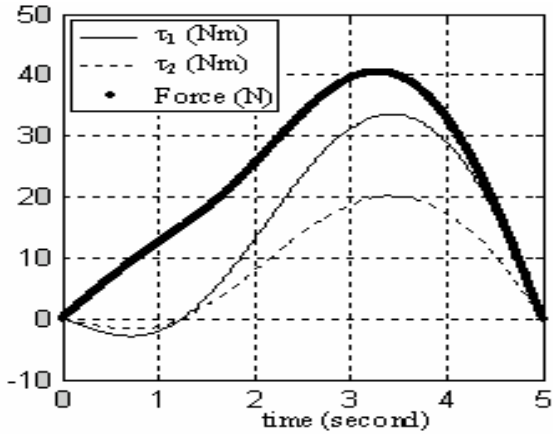
Fig. 4 Sinusoidal profile joint motion on stationary surface

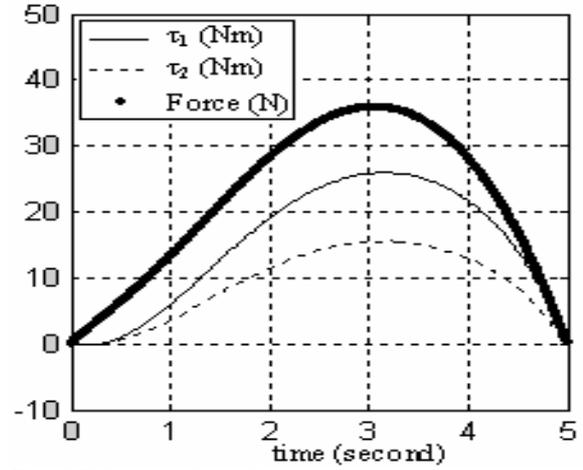
Fig. 6 Polynomial profile joint motion on stationary surface

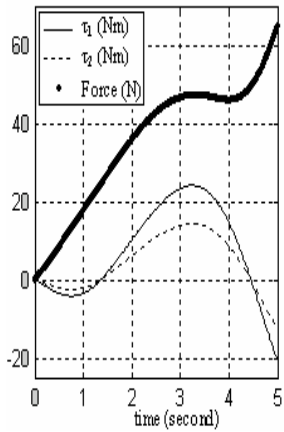 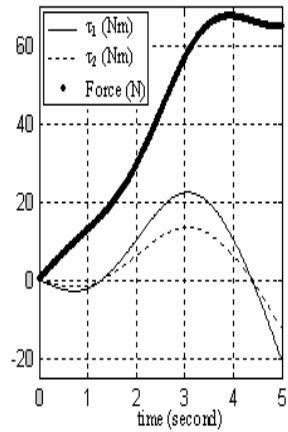
Fig. 5a Sine-Sine    Fig. 5b Gaussian-Sine

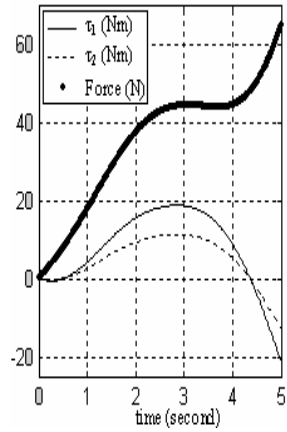 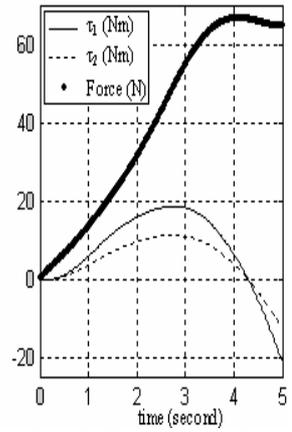
Fig. 7a Sine-Polynomial    Fig. 7b Gaussian-Poly.

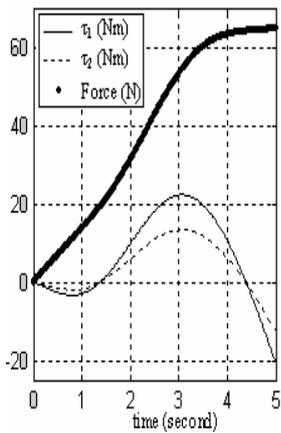 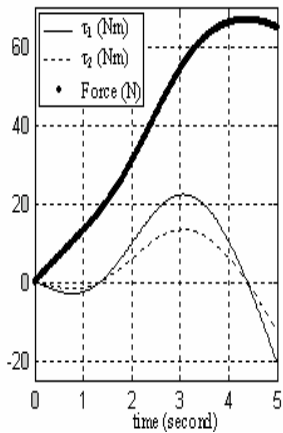
Fig. 5c Poly-Sine    Fig. 5d Parabolic-Sine

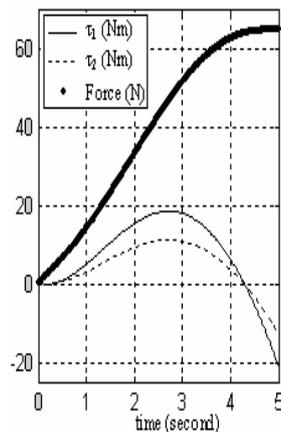 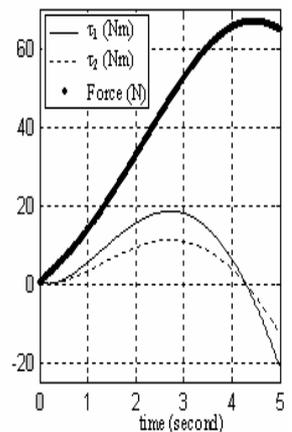
Fig. 7c Poly-Poly.    Fig. 7d Parabolic-Poly.

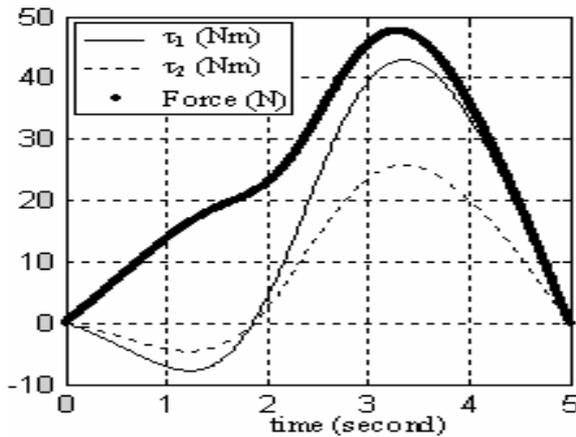

Fig. 8 Gaussian profile joint motion on stationary surface

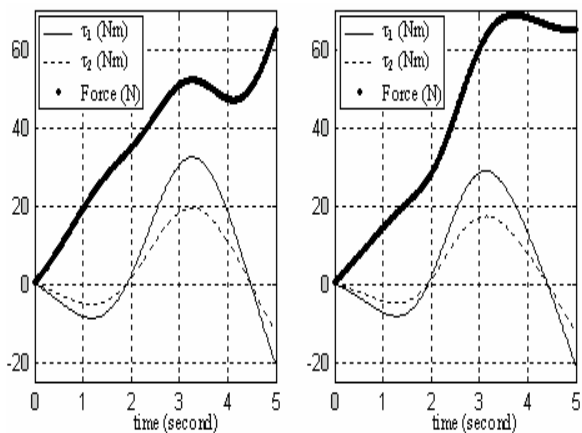

Fig. 9a Sine-Gaussian    Fig. 9b Gauss.-Gauss.

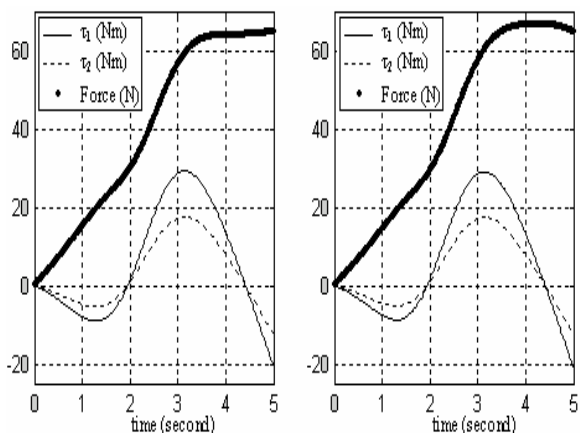

Fig. 9c Poly-Gauss.      Fig. 9d Parabolic-Gauss

## 5. Conclusion

In this paper, the effects of different angular velocity profiles for the constrained surface on the constrained force and the joint torques have been investigated. Simulation results show that system performance can also be affected due to the interaction between the joint motion and the angular motion of the constrained surface.

Therefore, for the better dynamic system performance, it is necessary to choose the motion profiles with judgment for both the surface rotation and the joint motion since there are two motions involved. However, knowledge of the time varying angle α is essential at every instant of surface rotation to obtain correct force measurements and thus joint torques.